# The Temporal Dynamics of Belief-based Updating of Epistemic Trust: Light at the End of the Tunnel?

**Momme von Sydow**[1,2] **(momme.von-sydow@urz.uni-münchen.de)**
**Christoph Merdes**[2] **(christoph.merdes@fau.de)**
**Ulrike Hahn**[1,3] **(u.hahn@bbk.ac.uk)**
[1] Ludwig-Maximilians-University of Munich (LMU), MCMP, Ludwigstr. 31, D-80539 München, Germany
[2] Friedrich Alexander University Erlangen-Nürnberg, ZiWiS, Bismarckstr. 8, 91054 Erlangen, Germany
[3] Birkbeck College London, Department of Psychological Science, Malet Street, London WC1E 7HX, U.K.

### Abstract

We start with the distinction of outcome- and belief-based Bayesian models of the sequential update of agents' beliefs and subjective reliability of sources (trust). We then focus on discussing the influential Bayesian model of belief-based trust update by Eric Olsson, which models dichotomic events and explicitly represents anti-reliability. After sketching some disastrous recent results for this perhaps most promising model of belief update, we show new simulation results for the temporal dynamics of learning belief with and without trust update and with and without communication. The results seem to shed at least a somewhat more positive light on the communicating-and-trust-updating agents. This may be a light at the end of the tunnel of belief-based models of trust updating, but the interpretation of the clear findings is much less clear.

## Introduction

Socially available information is an indispensable source of knowledge, from expert judgment to witness testimony. However, learning from partially reliable sources is fraught with problems. In particular, it is difficult to simultaneously update a belief in a proposition of interest and the reliability of a particular source providing pieces of evidence. Lacking certainty on either component, epistemic agents have to face such scenarios frequently. We focus on simple dichotomic hypotheses ($H = 0$ or $H = 1$), partial reliable evidence, and Bayesian agents when discussing the issue of the dynamics of sequential belief-based updating.

## Outcome-based vs. belief-based strategies of updating epistemic trust

As an introduction we first clarify our distinction of outcome-based and belief-based strategies of trust updating (mentioned in Hahn, von Sydow, Merdes, 2018). Although our main concern is with belief-based strategies, we will here sketch outcome-based strategies in slightly more detail and discuss a potential problem.

**Outcome-based strategies** A rather unproblematic method for updating epistemic trust seems to be outcome-based strategies, using observed frequencies of the co-occurrence of (apparent) evidence and ultimate outcomes. In such situations, the standard tools of frequentism or Bayesian inference seem applicable to represent rational degrees of reliability or 'epistemic trust'. Such strategies, however, assume that (all) agents can look behind the veil of phenomena (the evidence) and – at some point – have access to the truth (at least pragmatically, even if the thing-in-itself may never be known).

It seems, for instance, that we can come to judge quite accurately the reliability[1] of medical tests, such as a pregnancy test, only if we can square instances of test prediction with the eventual outcomes.

Technically, one needs an estimate for the originally unknown likelihood that the evidence of the test is positive ($D_1$='pregnant'), given the hypothesis is true ($H_1$ = truly pregnant), $P(D_1|H_1) = P(\text{'pregnant'} | \text{pregnant})$. This is the true positive rate, also called the *sensitivity* of a test. For a dichotomic outcome (with $H_2$ = not pregnant) it follows that $P(D_1|H_1) = 1 - P(D_2|H_1)$. Likewise one can infer the true negative rate, also called 'specificity' $P(D_2|H_2) = P(\text{'not pregnant'} | \text{not pregnant})$. More directly, one can obtain evidence for the positive predictive value, $P(\text{pregnant}|\text{'pregnant'})$, and negative predictive value, $P(\text{not pregnant}|\text{'not pregnant'})$. Let us assume for simplicity's sake that the reliability of the test remains stable over time. One possibility for assessing these conditional probabilities is the frequentist approach of counting the positive relative to all instances. Another possibility for assessing rational expected conditional probabilities and conditional probability distributions over a parameter Theta is a Bayesian, using priors, Bayes rule, conditionalization and a Beta-binomial updating (von Sydow, 2016).

Note, however, that any such outcome-based approach, although providing rational trust values, already assumes, by definition, 100% reliable final outcomes. On the other hand it excludes basing reliability estimates on one's beliefs and is strictly speaking not applicable if this 100% reliability is not ensured. Philosophically one may therefore argue that this involves a regress, even *ad infinitum*: how is our 100% trust in the final observation justified? One possibility is linked to a respectable tradition in philosophy advocating distinctions such as theoretical vocabulary *vs.* observational vocabulary, knowledge by descriptions *vs.* knowledge by acquaintance, or 'seeing as' *vs.* 'seeing that'.

---

[1] Reliability in our debate is defined as conditional probability or likelihood. For instance, the generative objective reliability is the conditional probability $P(\text{Test } H=1|\text{Truth } H=1)$, which is set equal to $P(\text{Test } H=0|\text{Truth } H=0)$, according to which the data is sampled. Note that this notion of reliability differs from other notions used in diagnostics (such as re-test reliability).

However, first, these distinctions and the 'Myth of the Given' (Sellars, 1956) have been challenged by philosophers and psychologists of perception, suggesting that observations on some level may be illusory, contextual or theory-laden. Second, even if one accepts such distinctions, one may counter that this is not helpful here: Even if one assumes basic, reliable sense data, this does not ensure that the 'observations' we are interested in here are 100% reliable. One cannot directly 'observe' that one is/was pregnant, but only a belly or later, a baby born. Our contextual knowledge plays a role to interpreting such 'observations'. It seems even questionable whether in principle one is ever entitled to be 100% sure (on contingent matters), particularly if one is a Bayesian. Another possibility might be that there are knowledge-based, theoretical or contextual justifications for assigning very high trust in observations (perhaps even involving inductive virtuous circles; von Sydow, 2006).

However, at least from a more pragmatic viewpoint, our distinction seems to make sense, since there is an obvious difference between a test whose reliability is not known, and a later obtained outcome that – by common standards – can be known. Nonetheless, we thought it is also adequate to raise the question whether our dichotomic distinction of outcome-based and belief-based trust updating has its problems and is, perhaps, too simplistic.

**Belief-based strategies** Even if there are clear cut outcome-based situations, many actual situations seem to involve and require *belief-based* strategies of trust updating. For instance, in the contexts of communication, reading, witness reports (and perhaps even measurement) we often obtain reports that are only partially reliable, without ourselves being able to assess their ultimate truth. In a belief-based strategy of trust-update, an agent nonetheless uses this evidence and revises not just her beliefs, but also the reliability of the respective source based on the match between evidence and current belief (Olsson, 2011, 2013; Bovens & Hartmann, 2003). For instance, if you tell me that the Earth is flat, this will make me think this is slightly more likely to be true, but it will also make me consider you less reliable than I had previously thought. (Incidentally, in a curved space the earth can in fact be considered flat.) Such a strategy seems intuitive, and there is empirical evidence supporting its use (Collins, Hahn, von Gerber, & Olsson, 2018).

## Olsson's Bayesian model of belief-based trust-update

We here concentrate on an influential Bayesian model of belief-based trust-update formulated by Eric Olsson (2011, 2013). It explicitly represents anti-reliability (see also Bovens & Hartmann, 2003, for an interesting alternative, cf. Merdes et al., subm.).

**The model** incorporates various parameters, such as a threshold for full belief, a probability of communication if this threshold is passed, etc. However, the following two equations are central. The first specifies the Bayesian update of the belief (credence or subjective probability) about a dichotomic hypothesis according to

$$P_{t+1}(h) = P_t(h|e) = \frac{E[\tau_t] * P_t(h)}{E[\tau_t] * P_t(h) + (1 - E[\tau_t]) * (1 - P_t(h))}$$

where E[tau] is subjective reliability, formalized as expected value of a reliability distribution. The reliability distribution is updated by:

$$\tau_{t+1}(x) = \tau(x|e) = \frac{x * P_t(h) + (1 - x) * (1 - P_t(h))}{E[\tau_t] * P_t(h) + (1 - E[\tau_t]) * (1 - P_t(h))} \tau_t(x)$$

Although various priors can be used, for the purpose of this paper a natural beta distribution is used, which is the conjugate standard distribution for binomial events.

### Previous Results - A disaster

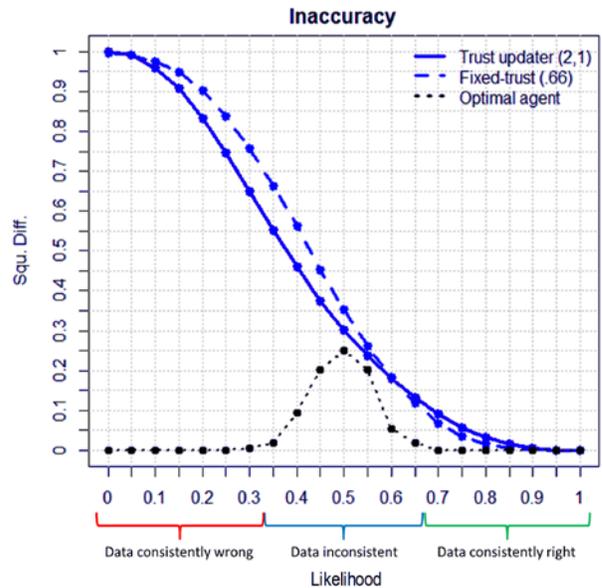

Figure 1: Mean (in)accuracy (squared error) of agent's belief averaged over 1000 runs for each objective likelihood and base rate of *H* being true. The graph is collapsed over base rates, since they had no influence here. Each run involved 10 pieces of evidence. The lines compare trust updater and fixed-trust agent and set them against the accuracy-score of an optimal agent who knows the true likelihood (from Hahn et al., 2018).

For several reasons, one may argue that although this model uses Bayes theorem it is concerned with *naïve* **Bayesian agents** only (and not necessarily optimal Bayesians). This is linked to issues of independence of sources and using the model sequentially (Hahn et al., 2018; Merdes et al., subm.). We nonetheless treat it as a possibly reasonable "Bayesian heuristics" whose normative merits must be explored empirically.

Recent simulations (Hahn et al., 2018; Hahn et al., 2019, subm.; Merdes et al., subm.) showed some disastrous results, calling into doubt that belief-based updating of source epistemic trust is a normative, optimal or even good strategy.

Hahn et al. (2018) investigated the Olsson model, looking at effects of trust-update on individual isolated agents. The Olsson model indeed also involves updating the reliability of data sources. Due to the feedback circle between belief and trust in successive updating (and in the network context also due to potential dependencies between communication-sources), learning is order-dependent. Based on agent-based modelling and simulations comparing trust-updating with fixed-trust agents, Figure 1 presents one comparison of them. The graph shows the inadequacy of the agent's belief. In the simulations the isolated agents began with a particular subjective reliability or trust ($p_{sub}$=.66), facing different objective reliabilities $p_{obj}$. The results show not only that the trust-updating (despite formally representing anti-reliability) clearly does not solve the problem of anti-reliable data (or the 'problem of a Cartesian Demon'), but it also shows that trust-updaters and fixed-trust agents behave quite similarly even with a disadvantage to the average beliefs when $p_{obj} \gg .5$ (actually for absolute differences for $p_{obj} > .5$). Surprisingly, updaters here fare even worse than fixed-trust agents.

In further simulations without networks of updating or fixed-trust agents, similar results have been found (Hahn et al., subm.; cf. also Hahn et al., 2018b).

## Light at the end of the tunnel?

Before proceeding to our new result, we briefly wish to point out another finding that could be interpreted somewhat more positively for belief-based updating.

### Prior knowledge

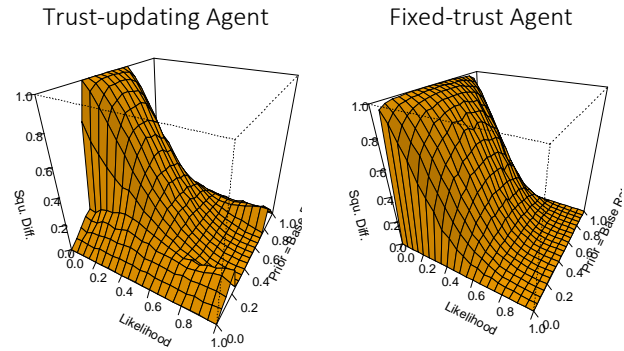

Figure 2: Mean (in)adequacy (the squared difference between beliefs and the actual status of hypothesis *H* (false = 0, true = 1) for either trust-updating or fixed-trust agents. The agents had prior knowledge of Prior = Base rates (from Hahn et al.,2018, and one panel of Fig. 6)

A further result of our previous simulations (Hahn et al, 2008) is that if the agents were endowed with *some* prior knowledge, this led to somewhat better results. It has also claimed more generally, with reference to the fundamental problem of induction, that induction, perhaps paradoxically, always requires to be a knowledge-based endeavour (von Sydow, 2006). However, Figure 2 shows the (in)adequacy of the final individual average beliefs for trust-updating and fixed-trust agents given some prior knowledge (prior = base rates). We cannot discuss the results in detail here, but it is apparent that for huge areas of the parameter space the error (high regions in the landscape plot) remains high with both strategies. However, the updater here turned out somewhat better, with an advantage relative to the fixed-trust agent. Additionally, an adequacy-improvement was found with regard to the trust-values themselves. Although the results for the updater are not good enough to elicit enthusiasm, they suggest advantages in situations with prior knowledge. Future work has to show whether trust-updating may have enhancing effects when, for instance in more complex belief networks, belief-based updating is mixed with fixed-trust updating.

## Dynamics – our new contribution

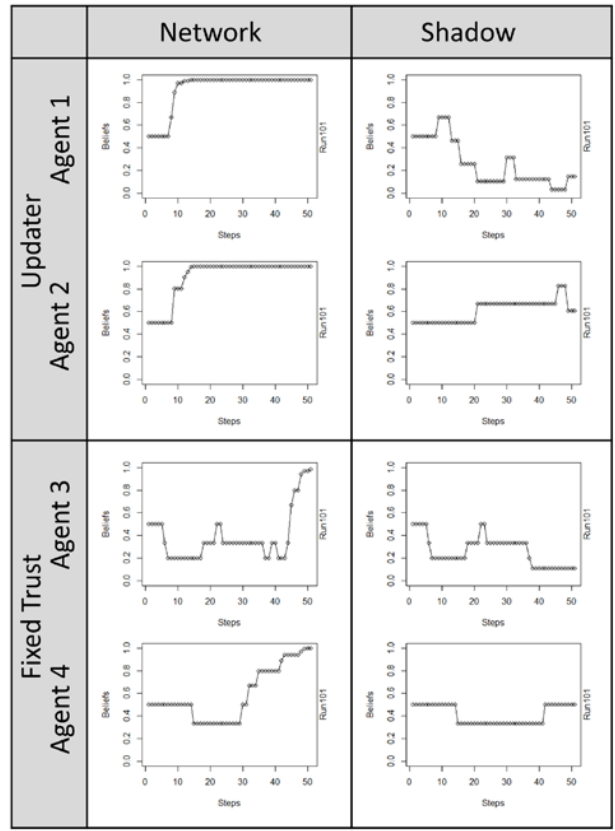

Figure 3: Belief in true hypothesis over 50 modeling steps of two agents in a run with objective reliability *p* = .66

In the remainder of this paper we focus on presenting our new results on the temporal dynamics. We investigate advantages and disadvantages of communicating vs. non-communicating agents and belief-based updating vs. fixed-trust agents over time. In this presentation we concentrate on exemplary parameter values.[2] Given these values, the chart panels presented begin with looking at quite specific examples, to more general findings. Crucially, the hypothesis here is always set to be true ($H=1$, and $P_{obj}(H) = 1$). We show belief plots only, but they can be interpreted as adequacy plots, given that the differences from one represent absolute (not squared) inadequacy.

---

[2] For simplicity and reduction of free parameters, we follow Olsson, using symmetric objective and subjective likelihood (=

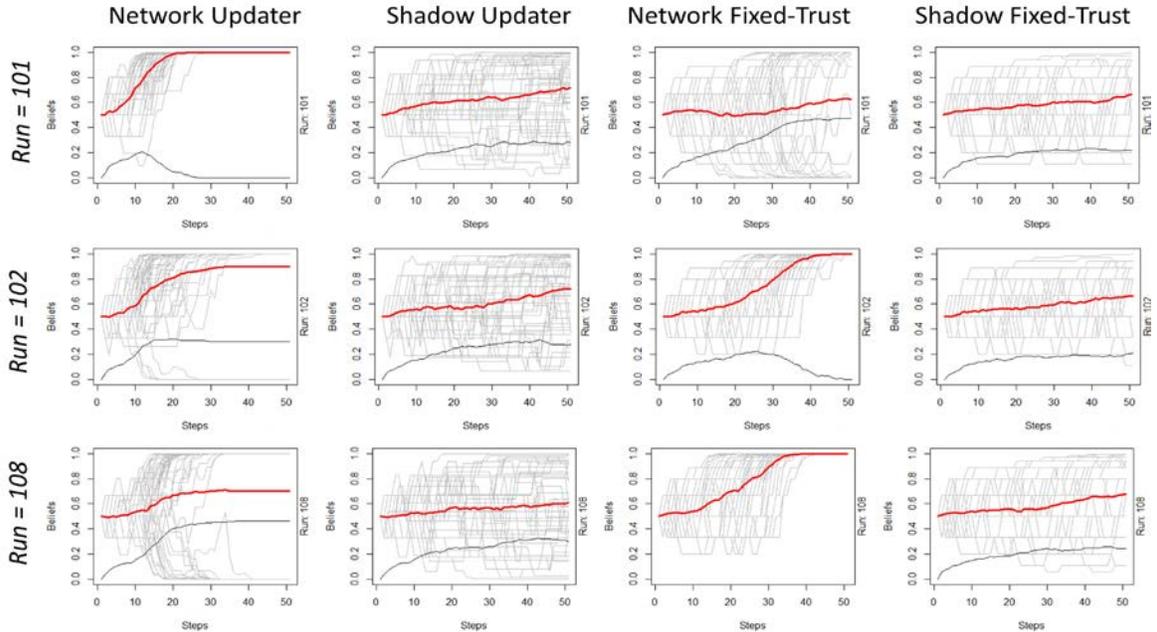

Figure 4: Three examples (three runs) of belief dynamics of a network for an objective reliability of .66 of the four kinds of updaters (Network Updater, Network Shadow, Fixed-Trust Agent, Fixed-Trust Shadow). Each graph shows the belief change of the 50 agents in the network (each is represented by a gray line), a red line shows mean beliefs and a black line the variance of agents' beliefs. The evidence is matched only for pairs of network and shadow agents (e.g. for Updaters).

Figure 3 provides a first impression of how belief is updated over 50 rounds for two random agents out of a random run with 50 agents. The agents are either communicating ('network' agents) or shadow agents. They either have a fixed subjective reliability/likelihood (= trust) or are belief-based updates of trust. It is apparent that the fixed-trust agents only 'move' through specific belief levels. This follows from the fixed (and symmetric) likelihood function, and it corresponds to the fact that here, without belief-based trust updating, learning is not order-dependent. Obtaining first evidence $E$ (favouring $H$) and then $\neg E$ (favouring $\neg H$) leads back to the identical belief level, and these plateaus of belief should not change even after time.[3]

In contrast, trust-updating agents do not get back to their previous plateaus. If we compare network and shadow agents, who actually received identical data from the world but either communicated or not, the non-communicating agents seem less decided. Some examples here have by chance actually partly obtained more evidence counter to rather than in favour of the hypothesis. In contrast, for the communicating agents, the evidence obtained from other agents helped develop a resolved belief in the true hypothesis $H$.

Although the results of Figure 3 are helpful for a first impression, Figure 4 generalizes them in showing results for all 50 agents of a run (gray lines) and all four investigated kinds of updating in single graphs. Additionally, the red lines show the means, and the black lines the variances between agents. The three examples suggest the following: The shadow fixed-trust updating for low reliability ($p = .66$) leads to beliefs with great variance. The updating fixed-trust agent's belief shows an even larger between-agent variation, presumably due to self-enforcing dynamics of belief-based updating. Communication here clearly seems to improve the situation by amplifying individual's over-weak signals. Here the variance often diminishes, although not in all examples. The network updating strategy in fact brings back agents starting with a dynamic in the wrong direction (Run 101), but it also seems to show partial (Run 102) or complete and stable polarization within a run (Run 103).

---

reliability) functions with $P(H|E) = P(\neg H|\neg E)$. Further used parameters were belief prior, $P_{sub\_bel}(H) = .5$; trust prior, $P_{sub\_rel}(Sources) = .66$ (for fixed trust and, for the updater, a Beta 2 1 distribution); threshold of assertation = 0.8; probability of communication = .25; and probability of occurrence data = .1. The topology was a standard small world network (50 agents, 2 neighbours, rewiring probability of .2). We simulated 50 steps of updating. We also used a high integration resolution (1000) and an improved algorithm (which did not seem to matter very much).

[3] This seems to follow from a standard Bayesian updating. However, one might consider extending the Olsson model formalizing belief also as a second-order probability distribution over probabi-

lities (e.g., using a Beta distribution). Then (without a kind of forgetting function) updating movements would diminish with the accumulation of data.

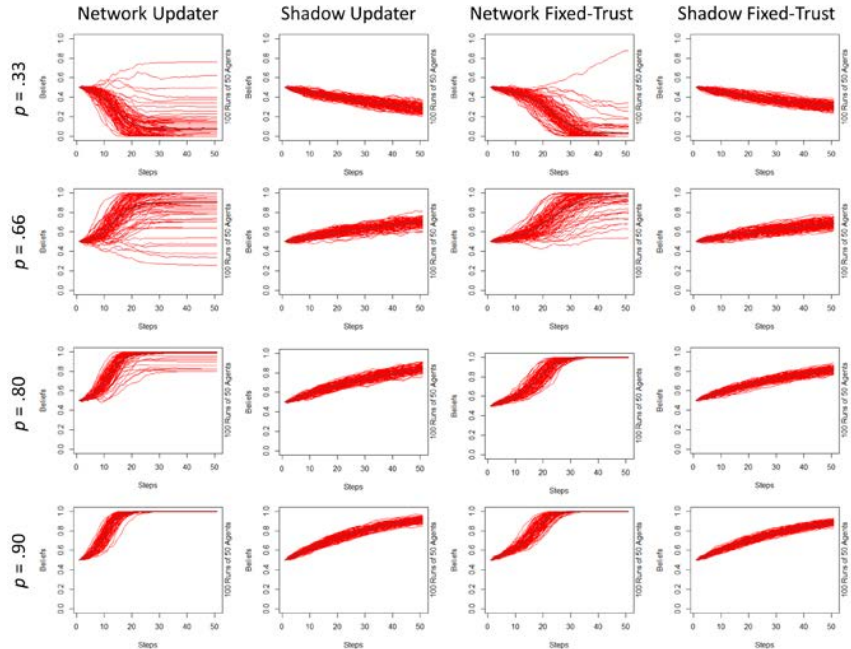

Figure 5: For each of the four investigated levels of objective reliability ($p = .33, .66, .80, .90$) we see the results for the four kinds of agents (Network Updater, Network Shadow, Fixed-Trust Agent, Fixed-Trust Shadow) who obtained the same data. In each chart we see the 50 agents' mean belief for 100 runs (red lines). Additionally, the purple lines show the runs' mean of these agents' mean beliefs.

To test the generality of these results, we simulated 100 model runs over 50 steps with these 50 agents—once with the updaters and once with the fixed trust agents (resulting in 500,000 agent data sets). Figure 4 gives the results. The red lines again show the mean beliefs of each run, now with 50 such lines in a graph. Additionally, their mean (the run's mean of mean agent's beliefs) is shown in purple. Figure 4 provides more general results for the reliabilities $p = .33$, .66, .80, and .90 (involving 2,000,000 agent results). It is apparent that the mean beliefs shown in Figure 5 for $p = .33$ and $p = .66$ is symmetrically reflected at an $x$-axis at $y = .5$. Figure 5 does not show within group/run variation (Fig. 4) but rather the variation *between* the group/run means. It becomes apparent that this variation at least is clearly much lower for shadow agents than for (communicating) network agents, particularly for non-extreme levels of reliability ($p = .33$ and $p = .66$). Although the communication normally reduces within-group variance (Fig, 4), Figure 5 shows that it increases the variance between groups in different runs—particularly if paired with belief-based updating.

To compare just the means of means (purple lines) in the updating strategies over time, we also plotted beside each other in Figure 6 for each of the four reliability values.

On this level of aggregation of means of mean beliefs, first show that they go in the correct direction as long as the reliability is $p > .5$, and in a negative direction with an anti-reliable data $p < .5$. But it has previously been pointed out that, even though formally modelling anti-reliability, the Olsson's Bayesian model of belief-based trust-update obviously does not address the problem of anti-reliability (the problem of the 'Cartesian Demon'), at least not without any prior knowledge present in this epistemic system (Hahn et al., 2018, subm.). The graphs also show the symmetry of partially reliable and partially unreliable sources.

Apart from this, if we exclude anti-reliable sources and only look at $p = .66, .80, .90$, both kinds of (non-communicating) shadow agents clearly seem less resolved than the networking agents (and highly similar to each other with a small advantage for the updater even). Interestingly, for the communicating agents there is a pronounced cross-over of graphs of updater and fixed-trust agent: At the beginning, the updater clearly is in advantage, then the fixed-trust agent catches up and overtakes the updater (the latter visibly only for $p = .66$). If we look back at Figure 4, it becomes apparent that the updating network-agent pays a price for quickly approaching a higher belief in true $H$ (given one is on the right side of reliability).

The price seems a higher number of runs where the mean belief does not approach 1, but seems to approach another attractor. This results from trust-updating stabilizing polarizations (cf. Fig. 3) where groups have opposed beliefs, by potentially turning counter-information into anti-reliable information in favor of one's own hypothesis. This occurs here not based on motivational concerns, but actually based on 'neural' belief-based updating of reliability alone. Moreover, it appears that although the shadow agents plausibly fare less well over time than the communicators, they will presumably at some point overtake the communicating agents. We have not simulated this up to now, but at least for the shadow fixed-trust agent this is theoretically clear. They are very slow due to not amplifying the signal--either by using information of other agents or their own beliefs to modify the reliability of sources. They only use the facts (with a predefined reliability), without being trapped in any feedback loop. For a partially positive reliable source (for $p_{obj} > .5$), a fixed-trust agent with $p_{sub} = .66$ will converge to $P(H) = 1$ in the long run. Given that network agents', and particularly network updaters', beliefs have stable attractors differing from 1, it follows that the isolated epistemic atom, will at least in the very long run outperform all refined others.

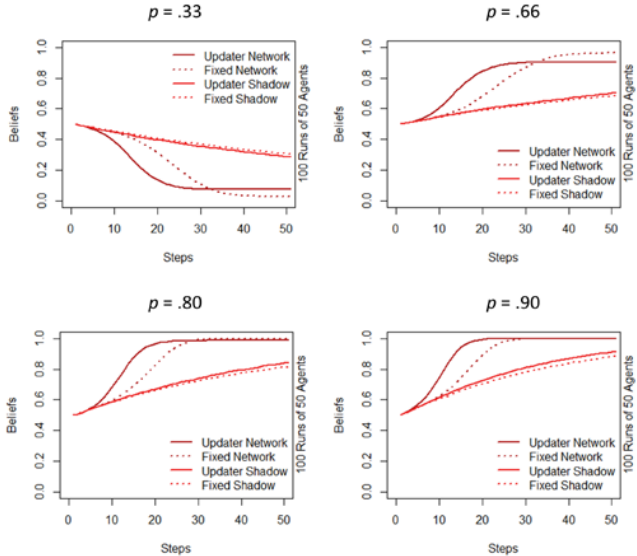

Figure 6. Mean of mean beliefs for the four updating strategies and four starting (or fixed) trust values ($p = .33$, .66, .80, .90) over time (again with $H$ always being true, and higher values thus displaying more accurate beliefs).

## General Discussion

Whereas our previous findings stressed the surprising similarity of belief-based trust updaters to fixed-trust agents in single-agent simulations (Hahn, von Sydow, & Merdes, 2018) and the positive role of communication (for most parameterizations) for average individual votes but not for majority votes (Hahn, Merdes, & von Sydow, 2019), the current findings aim to shed light on the temporal dynamics and suggest that communication as well as trust-updating (particularly if paired together) can have advantages in the short run but may pay for this by being disadvantaged in the long run. For instance, the most aggregated Figure 6 shows that if we look at the panel for reliability $p_{obj} = .66$ a switch over time of the resulting belief of the four updating strategies, which here corresponds to their order to be close to the truth (here $H$ is always set to 1). All agents began with the (underconfident) value of $P(H=1) = .5$. Then the order, say after 20 updating steps, was UpNe >> FiNe >> UpSh > FiSh, but in the long run it switched to FiSh > UpSh > FiNe > UpNe (cf. Figures 4 and 5). This seems – at least for the parameters and topology investigated – to hold for $p_{obj} > .5$. However, even here, for the subset of very high $p_{obj}$ values, the four strategies will not switch but only approach the same value 1, because no polarization takes place. For these $p_{obj}$ values, the beliefs of the Network Updaters all finally approach the true value of $P_{sub}(H = 1) = 1$ as well, but also much more quickly than fixed-trust agents.

Likewise, communication in the short run improves the confidence in a true hypothesis. However, the communicating (networkers) updaters' advantage of a quick initial convergence to the true value (as apparent for $p_{obj} = .66$) is only achieved by the price of polarization (Figures 4 and 5) and hence no complete convergence (Figure 6).

One may object that polarization need not be a disadvantage. In another research paradigm, using active sampling and an exploration exploitation trade-off, it has been convincingly shown that *transient* group diversity could keep up the diversity required to find underexplored optimal options (Zollman, 2007, 2010). However, the paradigm and parameters can matter (cf. Frey, & Šešelja, 2018), and we are here concerned with a different observation paradigm. The observation paradigm treats the agents as more passive than the active sampling paradigm (and reminiscent rather of classical than of operant conditioning). Even in the studied observation paradigm, agents are not fully passive due to weighting of external evidence by internal trust. In any case, in the observation-paradigm the polarizations are here shown to be normally non-transient – at least when simulated in our simple one-proposition networks (Figure 4, Network Updater, Run 108; Figure 5, Network Updater, horizontal lines not approximating 1).

To return to our main issue, polarizations here are non-transient and negative. This is a major reason for the fixed-trust agents' improving in the long run. Additionally, communicating seems helpful particularly for network agents, but will be matched or outperformed in the long run by non-communicating shadow-agents.

Although the analysis of the underlying dynamics and the switching result seems new and important for the observation paradigm, the interpretation of these results raises questions and requires further research. We in fact have not reached full consensus within our team on how to interpret the data. But this should not make them less interesting. A crucial question is whether to be as close to the truth as possible is always an advantage and whether an advantage in the short term can outweigh the advantage of being correct in the long term. This raises some fundamental issues that would require a more thorough treatment; here we can only briefly point out and discuss some open issues.

On the positive side, it seems prima facie that quickly achieving resolvedness can be an advantage if a belief in $H$ gets high (say, with H being 'a tiger is in the forest', and $H$ being true) and one quickly reaches a decision-threshold to retreat to safety as soon as possible.

A first objection may be that such considerations leave the realm of pure epistemology and require some potentially controversial assumptions about the epistemic or psychological constitution of the agent or the utilities relevant to the agent's decision making. This problem however, does not appear fundamental.

Another set of objections may concern how to weight (a) the utility of groups reaching a high mean belief in the truth in the short term against the price of some groups' going in the wrong direction in the long run; and (b) the non-transient heterogeneity even within groups.

There is a further fundamental problem of the advantages of achieving resolvedness quickly is that overconfidence seems never to be a good thing. This has at least two aspects: First, isn't the advantage of a high probability, $P_{sub}(H=1)$, given $H_{obj} = 1$, outweighed by a low probability of this proposition if $p_{obj}$ is actually low? This raises an important issue, but one may perhaps counter that more sources are presumably reliable ($p_{obj} > .5$) than anti-reliable ($p_{obj} < .5$). Therefore a weighted average model that takes this into account would put more weight on panels $p = .66$ and $p = .33$ (Fig. 6), and an advantage of the updater would remain. A second aspect is that it is not clear why we should not use even more radical values than those suggested by the Bayesian model. Given that an agent does not know whether his/her sources are truly independent (as assumed, but usually violated in actual social networks), and if the agent has no model of the dependencies (as assumed here), the Bayesian model is a Bayesian heuristic at best. Although the last point remains, rebuttal is also possible here. Agents started underconfident (with $P_{sub}(H) = .5$), and therefore the quick increase may be advantageous up to some point. We suggest simulating an ideal agent who obtains all information from all agents simultaneously as a comparison.

A final problem is that our graphs with $H_{obj} = 1$ implicitly used absolute-difference scores from the beliefs. However, the Brier score (squared difference) punishes variance and would here yield more problematic results for the short-term winners.

For all of these reasons the interpretation of our results concerning convergence-speed seems less clear than the true findings. We hope this sets the stage for further investigations and discussions. But let us return to the 'light at the end of the tunnel': Given that communication and also belief-based trust-updating are both prima facie and empirically plausible (Collins et al., 2018), isn't it also pragmatically justified to search for rational aspects of these mechanisms? Moreover, isn't it possible that potential speed-of-convergence advantage of belief-based updating combined with communication can positively interact with the second advantages of prior knowledge? Further discussions of these fundamental issues falls outside the scope of this paper, but it seems clear that our results suggest further avenues of research on agent-based modelling of belief-based trust update.

## Summary and Conclusion

Olsson has suggested a rational Bayesian model of sequential belief-based trust-update. Hahn et al., albeit partly having supported the model previously, have argued that it is a far from clear a priori that this is a rational model. Our simulations have shown that this belief-based updating led to partly devastating results. We here elaborated on the temporal dynamics of trust-updating and communication. In contrast to previous work, we here stressed the potentially remaining positive sides in reiterating some prior work on prior knowledge and, particularly, in showing that communicating updaters do seem to advantageous in more quickly arriving at more resolved beliefs. However, we also raised serious issues concerning the interpretation of these findings. Future avenues of research require further tests of boundary conditions of the dynamics and potential advantages of belief-based updating with prior knowledge (perhaps mixing belief-based and outcome-based updating). This work suggests more strongly than previous work that there may be 'light at the end of the tunnel' of negative findings on the normative status of belief-based trust-updating. However, at present it is not fully clear whether the light indicates the end of the tunnel or merely another oncoming train.